\documentclass[sigconf]{acmart}
\usepackage{threeparttable}
\usepackage{graphicx}
\usepackage{textcomp}
\usepackage{shortcuts}
\usepackage{framed}
\usepackage{xcolor}
\usepackage{url}
\usepackage{pifont}

\usepackage{breakurl}
\usepackage{enumitem}
\usepackage{multirow}

\usepackage[linesnumbered,titlenumbered,ruled,vlined,commentsnumbered]{algorithm2e}
\SetKwComment{Comment}{$\triangleright$\ }{}
\usepackage{comment}
\usepackage{enumitem}
\usepackage{dsfont}
\usepackage{booktabs}       % professional-quality tables
\usepackage{amsfonts}       % blackboard math symbols
\usepackage{nicefrac}       % compact symbols for 1/2, etc.
\usepackage{float}
\usepackage{microtype}      % microtypography
\PassOptionsToPackage{citecolor=royalblue,breaklinks,colorlinks,pagebackref}{hyperref}

\usepackage{appendix}

\usepackage{newfloat}
\usepackage{listings}
\DeclareCaptionStyle{ruled}{labelfont=normalfont,labelsep=colon,strut=off} % DO NOT CHANGE THIS
\floatstyle{ruled}
\newfloat{listing}{tb}{lst}{}
\floatname{listing}{Listing}

\definecolor{dg}{rgb}{0,0.694,0.298}
\definecolor{purple}{rgb}{0.4,0.176,0.569}
\definecolor{Gray}{gray}{0.6}
\definecolor{royalblue}{RGB}{65,105,225}

\makeatletter
\DeclareRobustCommand\onedot{\futurelet\@let@token\@onedot}
\def\@onedot{\ifx\@let@token.\else.\null\fi\xspace}
 
\def\ie{\emph{i.e}\onedot}

\makeatother

\AtBeginDocument{%
  \providecommand\BibTeX{{%
    \normalfont B\kern-0.5em{\scshape i\kern-0.25em b}\kern-0.8em\TeX}}}

\setcopyright{acmcopyright}
\copyrightyear{2018}
\acmYear{2018}
\acmDOI{XXXXXXX.XXXXXXX}

\acmConference[Conference acronym 'XX]{Make sure to enter the correct
  conference title from your rights confirmation emai}{June 03--05,
  2018}{Woodstock, NY}
\acmPrice{15.00}
\acmISBN{978-1-4503-XXXX-X/18/06}
\acmSubmissionID{185}

\begin{document}

\title{Seeing It Before It Happens: In-Generation NSFW Detection \\for Diffusion-Based Text-to-Image Models}

\author{Fan Yang}
\affiliation{%
  \institution{Huazhong University of Science and Technology}
  \city{Wuhan}
  \country{China}
}
% \email{m202472190@hust.edu.cn}

\author{Yihao Huang$^{\ast}$}
\affiliation{%
  \institution{National University of Singapore}
  \city{Singapore}
  \country{Singapore}
}
\thanks{*Corresponding authors. Yihao Huang and Kailong Wang}

\author{Jiayi Zhu}
\affiliation{%
  \institution{East China Normal University}
  \city{Shanghai}
  \country{China}
}

\author{Ling Shi}
\affiliation{%
  \institution{Nanyang Technological University}
  \city{Singapore}
  \country{Singapore}
}

\author{Geguang Pu}
\affiliation{%
  \institution{East China Normal University}
  \city{Shanghai}
  \country{China}
}

\author{Jin Song Dong}
\affiliation{%
  \institution{National University of Singapore}
  \city{Singapore}
  \country{Singapore}
}

\author{Kailong Wang$^{\ast}$}
\affiliation{%
  \institution{Huazhong University of Science and Technology}
  \city{Wuhan}
  \country{China}
}

\begin{abstract}
Diffusion-based text-to-image (T2I) models enable high-quality image generation but also pose significant risks of misuse, particularly in producing not-safe-for-work (NSFW) content. While prior detection methods have focused on filtering prompts before generation or moderating images afterward, the in-generation phase of diffusion models remains largely unexplored for NSFW detection. In this paper, we introduce In-Generation Detection (IGD), a simple yet effective approach that leverages the predicted noise during the diffusion process as an internal signal to identify NSFW content. This approach is motivated by preliminary findings suggesting that the predicted noise may capture semantic cues that differentiate NSFW from benign prompts, even when the prompts are adversarially crafted. Experiments conducted on seven NSFW categories show that IGD achieves an average detection accuracy of 92.45\% over naive and adversarial NSFW prompts, outperforming seven baseline methods.
\end{abstract}



\maketitle

\section{Introduction}
% 以下每一点是一个paragraph
%随着文本到图像生成模型的发展，其生成能力已显著提升。
%然而，这些模型也带来了滥用风险，例如绕过安全过滤器生成不当内容（如NSFW图像），对社会造成潜在危害。因此，开发有效的NSFW检测方法成为一个关键问题。
%现有的NSFW检测方法主要集中在生成前的文本分析（pre-detection）和生成后的图像识别（post-detection）两个阶段。前者依赖于自然语言理解模型对prompt进行过滤，但容易被同义替换、拼写变形或隐晦表达绕过，且难以准确感知生成结果的潜在风险；后者虽然可以直接识别生成内容，但需要对完整图像进行分析，计算开销大、检测延迟高，且只能在不良内容已被生成后“事后拦截”，存在不可逆的风险。
%在本文中，我们提出了一种新颖的“生成过程内检测”（in-generation detection， IGD）方法，用于识别扩散模型生成潜在NSFW内容的早期信号。不同于传统的pre-detection（文本过滤）和post-detection（图像识别）策略，我们的方法直接作用于生成模型内部的中间变量，在图像尚未完全合成时，即可做出有效判断。具体而言，我们聚焦于扩散模型生成过程中的一个关键变量——预测噪声（predicted noise），它是模型在每个扩散步骤中预测当前图像残差的核心输出。通过对大量样本的分析，我们发现：当模型生成不同类型内容（如SFW与NSFW）时，其对应的预测噪声在特征空间中表现出稳定而显著的分布差异。这种可分性使得预测噪声成为一种天然的、无需额外标注的检测信号。基于这一观察，我们设计了一种轻量级的检测机制，利用生成中间阶段的预测噪声特征进行快速判别，无需等待图像完全生成。该机制不仅有效提升了对NSFW内容的检测及时性，还具备较强的资源效率与攻击鲁棒性。实验结果表明，我们的方法在多个公开数据集上均取得了优异的检测性能，能够在不牺牲生成质量或引入显著计算开销的前提下，有效提升生成模型的安全性。

Text-to-image (T2I) models have seen rapid progress, largely driven by diffusion-based architectures. Models such as Stable Diffusion~\cite{Rombach_2022_CVPR}, DALL·E~\cite{ramesh2022hierarchical}, and SDXL~\cite{podell2023sdxl} can now generate high-quality, photorealistic images from natural language prompts, leading to widespread adoption in design, digital content creation, and virtual environments.
Yet, their growing capabilities also bring significant risks~\cite{liu2024jailbreak, zhang2024adversarial}. A primary concern regarding text-to-image (T2I) models is their potential misuse in generating illegal not-safe-for-work (NSFW) content, as opposed to safe-for-work (SFW) content that is legally acceptable. Some NSFW prompts are naive, being explicit and easily identifiable, while others are adversarially crafted to evade safety filters by subtly manipulating language or exploiting model vulnerabilities~\cite{yang2024sneakyprompt, yang2024mma, tsai2024ringabell, chin2023prompting4debugging}. These threats raise serious ethical and societal concerns, highlighting the need for more effective NSFW detection strategies.

%=================================================
\begin{figure}[tb]
\centering
        \includegraphics[width=\linewidth]{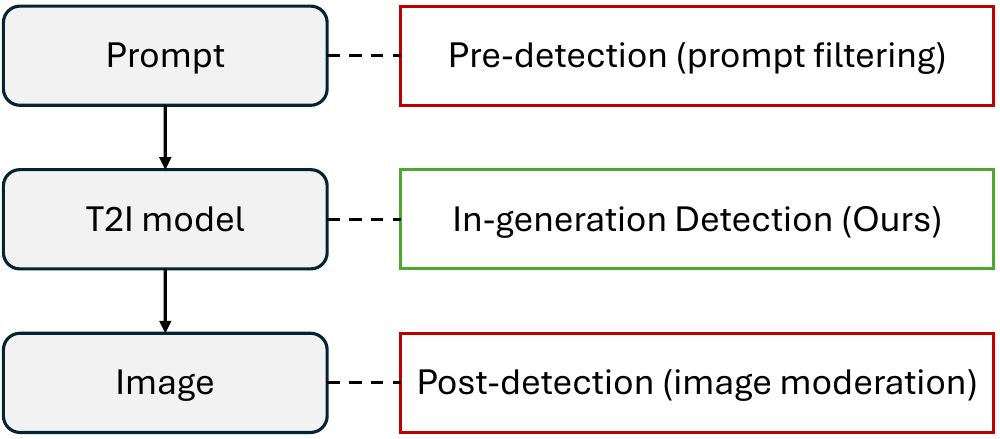}
	\caption{Overview of different NSFW detection types.}
	\label{fig:overview}
% \vspace{-6pt}
\end{figure}
% \vspace{-20pt}
%=================================================

Existing NSFW detection methods for T2I models can be broadly categorized into two types (see red boxes in Figure~\ref{fig:overview}): pre-detection, which analyzes prompts before generation, and post-detection, which evaluates the final image~\cite{liu2024jailbreak}. Pre-detection methods rely on lexical or classifier-based filters~\cite{li2023NSFWtext, Detoxify}, but are vulnerable to adversarially crafted prompts that use misspellings or vague language to obfuscate intent~\cite{yang2024mma, tsai2024ringabell}. Post-detection offers more accurate results~\cite{sd_safetychecker, openai2025moderation, aliyun_text_moderation, azure_ai_content_safety}, but introduces detection latency and more resource consumption~\cite{zhang2024adversarial}.

While existing methods primarily focus on pre-detection (prompt filtering) and post-detection (image moderation), the possibility of detecting NSFW content during the image generation process itself has, to the best of our knowledge, been largely overlooked. In diffusion-based T2I models, generation unfolds gradually over a sequence of denoising steps, offering a rich intermediate space that could be monitored in real time. However, this in-generation phase remains underexplored in the context of NSFW detection. 

To this end, we propose In-Generation Detection (IGD), a simple yet effective method for identifying NSFW intent during the image generation process in diffusion models. Specifically, IGD monitors the predicted noise throughout the denoising steps, which reflects the evolving visual semantics of the prompt. Leveraging this signal, we train a lightweight classifier to detect whether the given prompt is intended to produce NSFW content. The module integrates seamlessly into the generation loop, enabling timely intervention before the image is fully synthesized. Our approach is motivated by empirical observations showing that the predicted noise itself serves as a highly discriminative feature for NSFW detection. It captures semantic differences between prompts intended for NSFW and SFW content. Notably, this property also holds for adversarial NSFW prompts crafted to bypass prompt filtering. Despite their obfuscated surface form, the predicted noise patterns of these prompts closely resemble those of naive NSFW prompts, as both share the underlying intent to generate NSFW content. In contrast, SFW prompts tend to produce distinguishable noise patterns, enabling IGD to separate both naive and adversarial NSFW prompts from benign ones during generation. To sum up, IGD is lightweight, easy to integrate into T2I models, enables early intervention, and remains robust against adversarial prompts, making it a practical and effective solution for improving the safety of diffusion-based T2I systems. Experiments on seven NSFW categories show that IGD achieves an average detection accuracy of 92.45\% across both naive and adversarial prompts, outperforming seven baseline methods.

% contribution
In summary, our key contributions are:
\begin{itemize}[itemsep=0pt,topsep=0pt,parsep=0pt]
\item To the best of our knowledge, this is the first NSFW detection method that operates \textit{in-generation} by leveraging intermediate representations from diffusion-based T2I models, introducing a third paradigm beyond prompt filtering and image moderation.
\item Motivated by the observation that predicted noise in diffusion models encodes semantically discriminative patterns, we propose IGD, a simple yet effective method that leverages this signal for in-generation detection.
\item Experiments on seven NSFW categories show that IGD outperforms seven baselines and remains effective against adversarial prompts from five attack methods, demonstrating strong effectiveness and practicality.
\end{itemize}

\section{Related Work}
\subsection{NSFW Generation}
Prior research on generating NSFW content with text-to-image models can be broadly categorized into two approaches. The first involves collecting explicit prompts from online forums or NSFW communities, such as the I2P dataset~\cite{schramowski2023safei2p}, which contain direct descriptions of NSFW content without attempting to bypass safety mechanisms, termed \textbf{naive NSFW prompt}. The second focuses on adversarial prompting, where seemingly benign inputs are crafted to evade safety filters while still triggering NSFW outputs, termed \textbf{adversarial NSFW prompt}. Notable examples include SneakyPrompt~\cite{yang2024sneakyprompt}, which uses reinforcement learning to introduce subtle, filter-bypassing perturbations. MMA-Diffusion~\cite{yang2024mma} generates multimodal adversarial noise to mislead both text and image encoders. Ring-A-Bell~\cite{tsai2024ringabell} embeds semantic residues of removed concepts into innocuous prompts. P4D~\cite{chin2023prompting4debugging} employs automated red-teaming to uncover prompts that reveal safety vulnerabilities. DiffZOO~\cite{dang2024diffzoo} goes further by enabling purely query-based black-box attacks on diffusion models, approximating gradients via zeroth-order optimization to craft attack prompts that bypass content filters without any model internal knowledge.

% NSFW detection
\subsection{NSFW Detection}
Most existing NSFW detection approaches in T2I models operate at two stages: \textbf{pre-detection} (prompt-level) and \textbf{post-detection} (image-level).

Pre-detection methods analyze the input prompt to assess potential risks before image generation. Basic approaches use keyword-based filtering, while more advanced methods such as NSFW-text-classifier~\cite{li2023NSFWtext} and Detoxify~\cite{Detoxify} employ classifiers trained on large-scale corpora. However, these methods are vulnerable to adversarial prompts that obfuscate NSFW intent through synonym substitution, Unicode manipulation, or incoherent phrasing. Although GuardT2I~\cite{yang2024guardt2i} and Latent Guard~\cite{liu2024latentguard} leverage prompt embeddings for improved robustness, these embeddings themselves remain sensitive to surface-level perturbations.

Post-detection methods, in contrast, evaluate the generated image to determine whether it contains NSFW content. The Safety Checker~\cite{sd_safetychecker}, widely used in Stable Diffusion pipelines, applies CLIP-based classifiers to identify sensitive visual elements. Additionally, moderation services such as the OpenAI Moderation API~\cite{openai2025moderation}, Aliyun Text Moderation~\cite{aliyun_text_moderation}, and Azure AI Content Safety~\cite{azure_ai_content_safety} offer multimodal inspection across text and image modalities, covering a wide range of harmful content categories. While post-detection methods can be effective, they incur latency and act only after image generation. Although the image may not be displayed, the process still consumes significant computational resources and time before unsafe content is detected and discarded.

\section{Preliminary and Motivation}
% 文生图模型的基本结构和运行机理。主要介绍diffusion，参考以下内容做修改，要讲明predicted noise是怎么来的
\subsection{Preliminary}
\textbf{Diffusion models} are a type of generative model \cite{ho2020denoising} that decomposes the data generation process into two complementary stages: a forward process and a reverse process. In the forward process, noise is gradually added to the input image, transforming the original data distribution into a standard Gaussian noise. Conversely, the reverse process is trained to recover the original data from pure noise by learning to invert the corruption process.

Given an input image latent $x^0$, the forward process perturbs the data using a predefined noise schedule $\{\beta^t: \beta^t \in (0,1)\}_{t=1}^T$, which controls the magnitude of noise added over $T$ steps. This results in a sequence of noisy latent variables $\{x^1, x^2, \dots, x^T\}$. At each timestep $t$, the noisy sample $x^t$ is generated as:
%======================
\begin{equation}
x^t = \sqrt{\bar{\alpha}^t} x^0 + \sqrt{1 - \bar{\alpha}^t} \, \epsilon,
\end{equation}
%======================
where $\alpha^t = 1 - \beta^t$, $\bar{\alpha}^t = \prod_{s=1}^t \alpha^s$, and $\epsilon \sim \mathcal{N}(0, \mathbf{I})$ represents standard Gaussian noise.

The reverse process aims to denoise $x^{t+1}$ to obtain a less noisy $x^t$ by estimating the noise component $\epsilon$ using a neural network $\epsilon_\theta(x^{t+1}, t)$. The model is trained to minimize the $\ell_2$ distance between the true noise and the predicted noise:
%======================
\begin{equation}
\mathcal{L}_{\text{uncondition}} = \mathbb{E}_{x_0, t, \epsilon \sim \mathcal{N}(0,1)} \left\| \epsilon - \epsilon_\theta(x^{t+1}, t) \right\|_2^2,
\end{equation}
%======================
where $t$ is uniformly sampled from $\{1, \dots, T\}$.

In contrast to unconditional diffusion models, conditional (prompt-based) diffusion models guide the generation process using an additional condition or prompt $c$. This enables the model to produce photorealistic outputs that are semantically aligned with the given text prompt or concept. The training objective is then extended as:
%======================
\begin{equation}
\mathcal{L}_{\text{cond}} = \mathbb{E}_{x_0, t, c, \epsilon \sim \mathcal{N}(0,1)} \left\| \epsilon - \epsilon_\theta(x^{t+1}, t, c) \right\|_2^2.
\label{eq:condition_loss}
\end{equation}
%======================
% need an inference formula

%=================================================
\begin{figure}[t]
\centering
        \includegraphics[width=\linewidth]{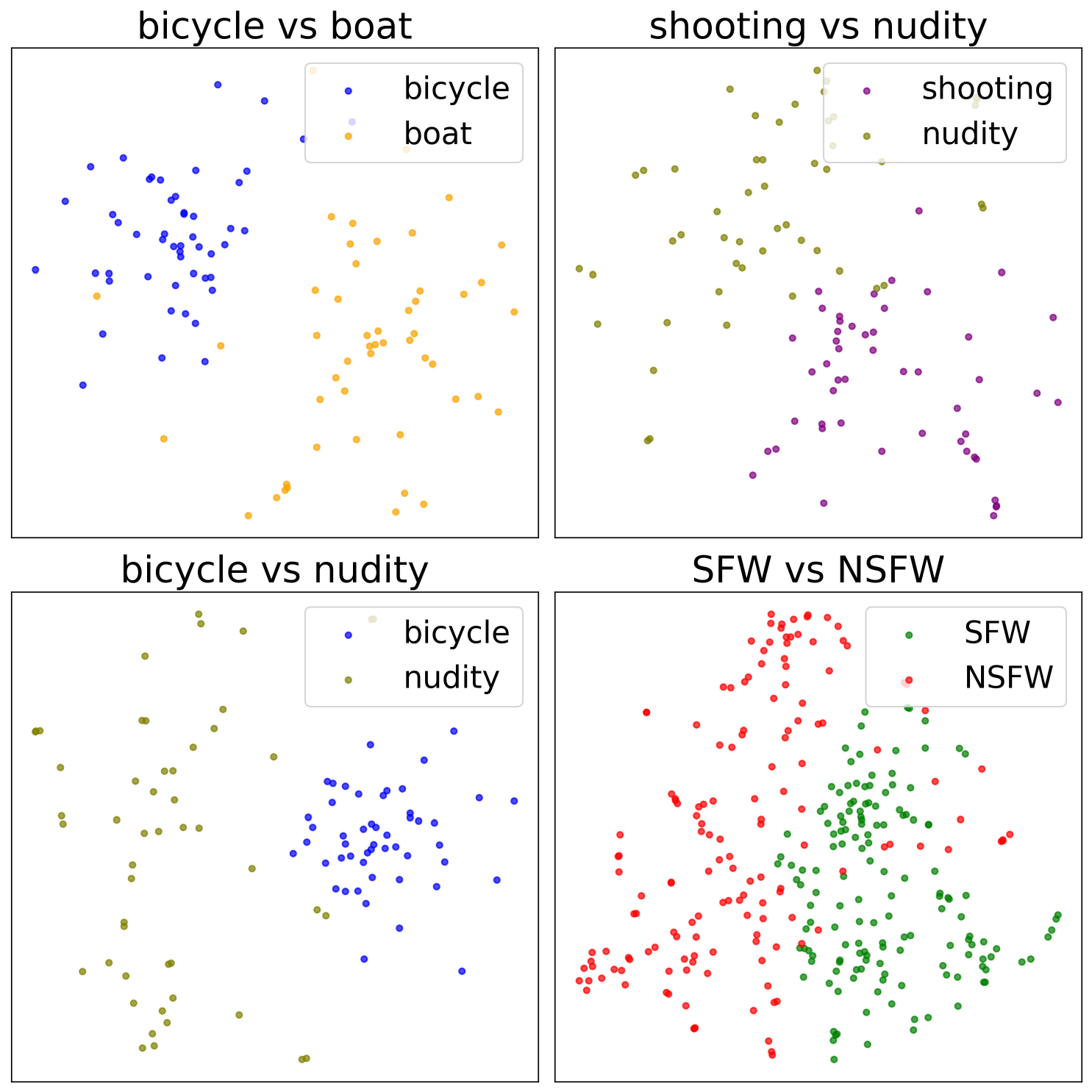}
	\caption{t-SNE visualizations of predicted noise $\epsilon_t$ across four representative category pairs. 
    % Top left: SFW vs. SFW (bicycle vs. boat), Top right: NSFW vs. NSFW (shooting vs. nudity). Bottom left: SFW vs. NSFW (bicycle vs. nudity), Bottom right: aggregated SFW vs. NSFW classes.
    }
	\label{fig:tsne2x2}
% \vspace{-6pt}
\end{figure}
% \vspace{-20pt}
%=================================================
%=================================================
\begin{figure}[t]
\centering
        \includegraphics[width=\linewidth]{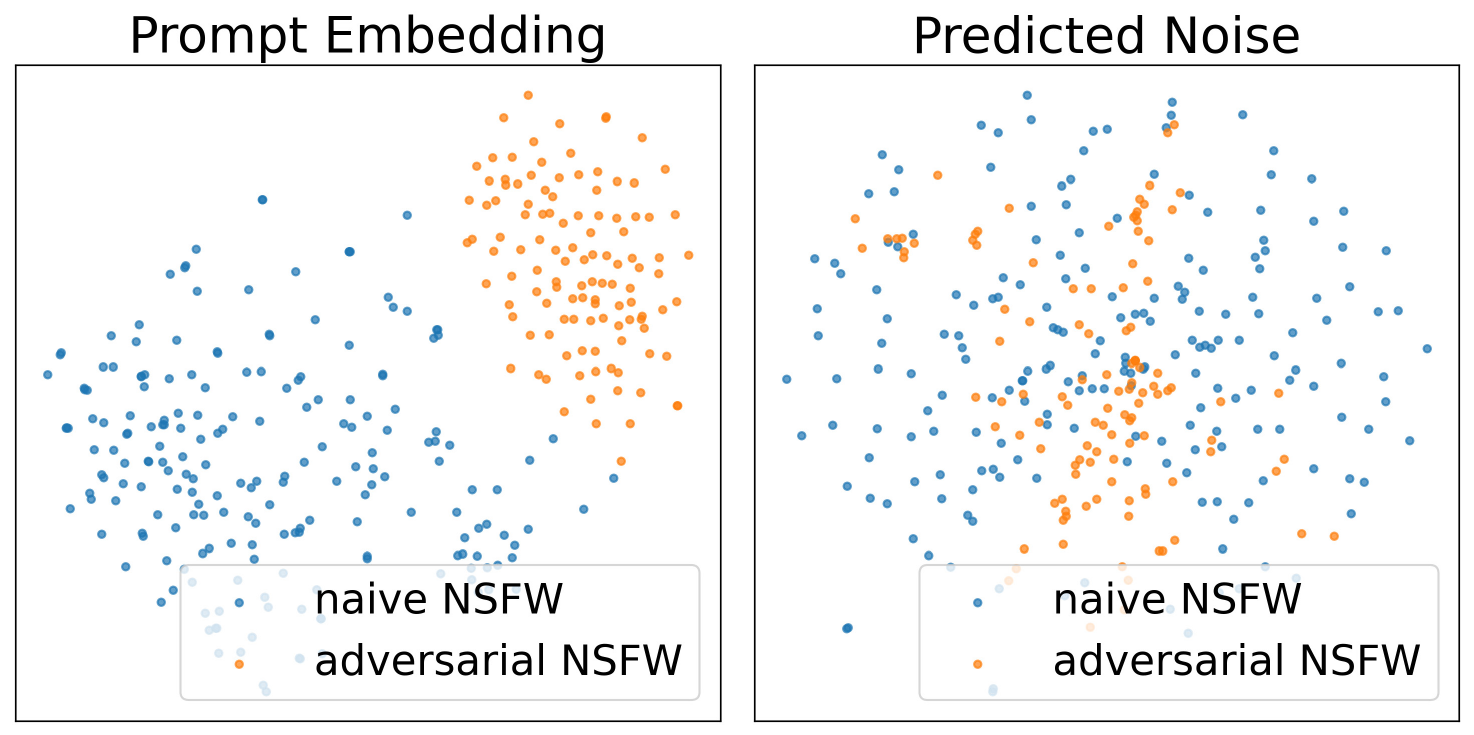}
	\caption{Comparing prompt embedding and predicted noise features between naive and adversarial NSFW prompt.}
	\label{fig:embed_vs_noise}
% \vspace{-6pt}
\end{figure}
%=================================================

\subsection{Motivation} 
According to Eq.~\eqref{eq:condition_loss}, the reverse process of diffusion (i.e., the denoising procedure) predicts the noise at timestep $t$ as:
%======================
\begin{equation}
\epsilon_t = \epsilon_\theta(x^{t+1}, t, c),
\label{eq:predicted_noise}
\end{equation}
%======================
where $x^{t+1}$ is the current noisy latent, $t$ is the diffusion timestep, and $c$ is the embedding of the text condition. Since the model is trained to generate images conditioned on $c$ (the embedding of the input prompt), the predicted noise $\epsilon_t$ naturally becomes a condition-dependent variable that reflects the semantic intent of the prompt and implicitly encodes information about the generated content. Consequently, it has the potential to serve as an effective feature for distinguishing the intent of NSFW and SFW prompts. \textbf{However, the potential of predicted noise in discriminative tasks, particularly NSFW detection, remains unexplored.}

%=================================================
\begin{figure*}[tb]
\centering
        \includegraphics[width=\linewidth]{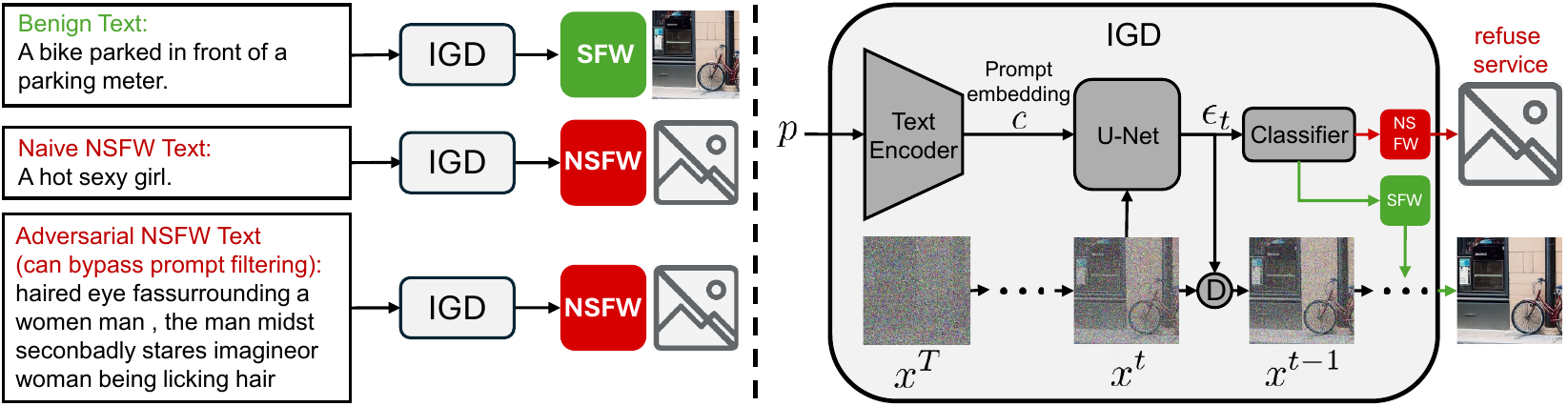}
	\caption{The overall framework of IGD.}
	\label{fig:framework}
% \vspace{-6pt}
\end{figure*}
%=================================================

\noindent\textbf{Predicted noise reflects separable generation intents for NSFW and SFW prompts.} To explore whether the predicted noise $\epsilon_t$ encodes meaningful semantic information, we conduct t-SNE~\cite{maaten2008visualizing} visualizations using $\epsilon_t$ extracted by Stable Diffusion v1.5~\cite{stablediffusion-v1-5} from prompts associated with different semantic labels. To be specific, we randomly select three SFW categories, including \textit{bicycle}, \textit{boat}, and \textit{train}, as well as three NSFW categories, including \textit{bloody}, \textit{nudity}, and \textit{shooting}. For each category, we use 50 different input prompts and analyze the distribution of the corresponding predicted noise $\epsilon_t$ at a random diffusion timestep. We also aggregate these into two broader classes, SFW and NSFW, to visualize their overall separation. 
As shown in Figure~\ref{fig:tsne2x2}, we construct a $2\times2$ panel illustrating comparisons between: \ding{182} two SFW categories (\textit{bicycle} vs. \textit{boat}); \ding{183} two NSFW categories (\textit{shooting} vs. \textit{nudity}); \ding{184} a direct SFW vs. NSFW category comparison (\textit{bicycle} vs. \textit{nudity}); and \ding{185} an aggregated comparison of three SFW categories (\textit{bicycle}, \textit{boat}, \textit{train}) and three NSFW categories (\textit{bloody}, \textit{nudity}, \textit{shooting}) (See Appendix~\ref{sec:a_tsne} for details). The resulting t-SNE plots show that the predicted noise forms semantically coherent clusters across different categories, with SFW and NSFW classes largely separated in embedding space. This indicates that $\epsilon_t$ encodes meaningful semantic structure before image synthesis is complete. While the analysis does not cover the full range of prompts, it offers important empirical insight: predicted noise captures discriminative patterns and holds strong potential as an in-generation signal for NSFW detection.

\noindent\textbf{Predicted noise as a signal for detecting adversarial NSFW prompts.} Adversarially crafted NSFW prompts can evade prompt filtering by subtly modifying surface text while preserving semantic intent, exposing a key weakness of pre-detection methods relying on prompt embeddings. In contrast, our in-generation approach IGD remains effective because both naive and adversarial NSFW prompts drive the model to generate unsafe content. This shared intent is reflected in the predicted noise during generation, which remains similar across prompt types despite textual obfuscation. Leveraging this convergence, IGD achieves greater robustness to adversarial prompts than pre-detection methods.

To illustrate this, we compare prompt embeddings and predicted noise features using representations from Stable Diffusion v1.5. Specifically, we use the \textit{sexual} category from the I2P dataset as naive NSFW prompts and generate adversarial variants using the Ring-A-Bell attack. For each prompt, we extract (1) the prompt embedding from the text encoder and (2) the predicted noise from the diffusion process. As shown in Figure~\ref{fig:embed_vs_noise}, the t-SNE visualization of prompt embeddings (left) shows a clear separation between naive and adversarial prompts. This reflects the success of adversarial attacks in manipulating surface text to shift the prompt embedding distribution. Since prompt-level classifiers typically rely on these embeddings to estimate NSFW likelihood, such separation allows adversarial prompts to evade detection despite preserving the original NSFW intent. In contrast, the predicted noise features (right) exhibit strong overlap between naive and adversarial prompts, suggesting that both lead to similar generative behavior during denoising. This convergence reveals that the underlying visual intent remains consistent, even if the textual form is obfuscated. As a result, a classifier trained solely on the predicted noise of naive NSFW prompts can still effectively detect adversarial ones. This highlights the robustness of predicted noise as a feature for in-generation NSFW detection, particularly under adversarial conditions.

\section{Method}
% Analysis on predicted noise，主要是noise是否能指明方向（第一组图），以及noise的t-sne能表明对类别的区分能力（第二组图）
\subsection{Problem Formulation}
The goal of NSFW detection is to determine whether a given input prompt $p$ will lead to the generation of inappropriate or sensitive visual content. Formally, let $p$ denote a text prompt, and let $y \in \{0, 1\}$ be the corresponding label, where $y = 1$ indicates that the image generated from $p$ contains NSFW content, and $y = 0$ otherwise. The task is to learn a function $f(p) \rightarrow y$ that predicts the safety label of the image conditioned on the prompt.
Although $y$ reflects the semantic class of the final generated image, the prediction can be based on various forms of input, such as the prompt itself, intermediate features during generation, or the final image. The output $y$ serves as a binary decision signal for content moderation.

% 方法的具体实现流程
\subsection{In-generation Detection Method}

We propose IGD, a simple yet effective in-generation NSFW detection method, as illustrated in Figure~\ref{fig:framework}. Unlike prior approaches that perform classification on the input prompt or output image, IGD leverages intermediate predicted noise from the diffusion process to identify unsafe content before image synthesis completes.

As illustrated in the right part of Figure~\ref{fig:framework}, given an input text prompt $p$, we first obtain its embedding $c$ using a text encoder $E_{\text{text}}$: $c = E_{\text{text}}(p)$. During the denoising process of a diffusion model (e.g., Stable Diffusion), the U-Net denoiser $\epsilon_\theta$ predicts the noise at each timestep $t$ as $\epsilon_t = \epsilon_\theta(x^{t+1}, t, c)$. We then attach a lightweight binary classifier $f_\phi(\cdot)$ to the predicted noise $\epsilon_t$ and define the NSFW decision as:
\begin{equation}
y = f_\phi(\epsilon_t),
\label{eq:nsfw_classifier}
\end{equation}
where $y \in \{0, 1\}$ indicates whether the predicted image is classified as NSFW. If $y = 1$, the generation is terminated early to prevent the synthesis of unsafe image. Otherwise, the process continues as usual. As illustrated in the left part of Figure~\ref{fig:framework}, IGD effectively handles benign prompts, naive NSFW descriptions, and even adversarially obfuscated texts that bypass prompt-level filters.

This method offers several benefits. \ding{182} Compared to pre-detection (prompt filtering), IGD is more \textit{robust} to obfuscated prompts. While prompt-based classifiers rely on surface text that can be easily manipulated, IGD analyzes the predicted noise $\epsilon_t$, which reflects how the model internally interprets and visualizes the prompt. This makes it less sensitive to minor textual variations and better aligned with the actual generation intent. \ding{183} Compared to post-detection (image moderation), IGD enables \textit{early intervention} by analyzing internal generative signals before image synthesis completes. \ding{184} Finally, the classifier is lightweight, consisting of a small number of neural layers, and introduces negligible overhead to the generation process.

\section{Experiment}\label{sec:experiment}
\subsection{Experimental Setups}
\noindent\textbf{Datasets.}
We conducted experiments on two datasets to train our NSFW detector: \textbf{I2P}~\cite{schramowski2023safei2p} and \textbf{MSCOCO}~\cite{lin2014microsoft}. The I2P dataset contains 4,703 manually crafted NSFW prompts targeting T2I models, covering seven NSFW categories: self-harm, violence, shocking content, hate, harassment, sexual, and illegal activity. We sample 200 prompts from each category, resulting in 1,400 NSFW training examples. For clean data, we use the training split of COCO2014, which includes 123,287 images, each paired with five human-written captions. We extract the first caption from each image and randomly sample 1,400 clean prompts. In total, the training set consists of 2,800 samples, evenly balanced between NSFW and clean data.

\noindent\textbf{Baselines.} 
We compare our method with \textbf{seven} representative moderation tools, covering open-source models, commercial APIs, and T2I-specific defenses. \textbf{NSFW-text-classifier}~\cite{li2023NSFWtext} is a Hugging Face-hosted binary classifier for NSFW text detection. \textbf{Detoxify}~\cite{Detoxify} is a transformer-based toxicity detector trained on Jigsaw’s dataset. \textbf{OpenAI Moderation API}~\cite{markov2023holistic,openai2025moderation}, \textbf{Aliyun Text Moderation}~\cite{aliyun_text_moderation}, and \textbf{Azure AI Content Safety}~\cite{azure_ai_content_safety} are commercial moderation services that support text inputs. They detect a range of harmful content categories, including sexual, violent, political, and hate-related material, using advanced multimodal or multilingual models.
% \textbf{OpenAI Moderation API}~\cite{markov2023holistic,openai2025moderation} detects harmful content across modalities using the \texttt{omni-moderation-latest} model. \textbf{Aliyun Text Moderation}~\cite{aliyun_text_moderation} is a multilingual real-time API for detecting violent, political, and explicit content. \textbf{Azure AI Content Safety}~\cite{azure_ai_content_safety} is a multimodal moderation service that flags sexual, violent, and hate content. 
\textbf{Latent Guard}~\cite{liu2024latentguard} detects adversarial prompts in the latent space of T2I embeddings without retraining. \textbf{GuardT2I}~\cite{yang2024guardt2i} interprets prompt embeddings via a conditional language model to identify harmful intent while preserving generation quality.

\noindent\textbf{Target model.} 
Following the same setting as previous baselines Latent Guard~\cite{liu2024latentguard} and GuardT2I~\cite{yang2024guardt2i}, we adopt Stable Diffusion v1.5~\cite{Rombach_2022_CVPR,stablediffusion-v1-5}, a widely used open-source T2I model, as our target model for obtaining predicted noise. 

\noindent\textbf{Evaluation metrics.} 
Following prior works~\cite{liu2024latentguard,yang2024guardt2i}, we adopt three standard metrics: \textbf{Accuracy}, \textbf{AUROC}, and \textbf{FPR@TPR95}. Accuracy reflects the classification correctness. AUROC measures the model’s discriminative ability across thresholds. FPR@TPR95 evaluates robustness under high recall. Higher accuracy and AUROC, and lower FPR@TPR95, indicate better performance.

% Following the evaluation protocols of Latent Guard~\cite{liu2024latentguard} and GuardT2I~\cite{yang2024guardt2i}, we adopt three standard metrics for NSFW prompt detection: \textbf{AUROC}, \textbf{Accuracy}, and \textbf{FPR@TPR95}. \textbf{Accuracy} captures the overall proportion of correctly classified prompts, providing an intuitive measure of general performance. \textbf{AUROC} quantifies the model’s ability to distinguish between adversarial and clean prompts by measuring the trade-off between true and false positive rates across all thresholds. \textbf{FPR@TPR95} measures the false positive rate when the true positive rate is fixed at 95\%, assessing robustness under high-recall conditions—an important consideration in practical applications where false alarms are costly. Higher AUROC and accuracy, along with lower FPR@TPR95, indicate better detection performance.

\noindent\textbf{Implementation details.}
We use Stable Diffusion v1.5 as the target T2I model to extract the predicted noise as the feature for classification. The total number of inference steps of the T2I model is 50, and the timestep we used for predicted noise extraction is 5. We employ a straightforward 5-layer fully connected MLP as a binary classifier, trained using the size-unfolded predicted noise as input features. Importantly, this classifier is trained solely on the naive NSFW and clean prompts from our constructed training set, without access to adversarial or paraphrased examples. The model is optimized using the Adam optimizer with a learning rate of $1e^{-3}$ for 100 epochs. We conduct our experiments on an NVIDIA RTX 3090 GPU with 24GB of memory.

%==============================================
\begin{table}[tbp]
\centering
\caption{Compare with baselines.}
\resizebox{\linewidth}{!}{
\begin{tabular}{c|c|cc|c}
\toprule
\multirow{1}{*}{} & \multirow{1}{*}{Methods} & Naive & Adversarial  & Average\tabularnewline
\midrule
\multirow{8}{*}{Accuracy $\uparrow$} & NSFW-text-classifier & 58.81\% & 71.12\% & 64.97\%\tabularnewline
 & Detoxify & 50.54\% & 56.38\% & 53.46\%\tabularnewline
 & OpenAI Moderation API & 57.50\% & 66.07\% & 61.78\%\tabularnewline
 & Aliyun Text Moderation & 52.78\% & 56.93\% & 54.86\%\tabularnewline
 & Azure AI Content Safety & 56.96\% & 72.77\% & 64.86\%\tabularnewline
 & Latent Guard & 57.26\% & 61.29\% & 59.28\%\tabularnewline
 & GuardT2I & 51.70\% & 65.20\% & 58.45\%\tabularnewline
 & IGD (Ours) & \textbf{90.96\%} & \textbf{93.94\%} & \textbf{92.45\%}\tabularnewline
\midrule
\multirow{8}{*}{AUROC $\uparrow$} & NSFW-text-classifier & 58.65\% & 62.76\% & 60.71\%\tabularnewline
 & Detoxify & 56.60\% & 71.56\% & 64.08\%\tabularnewline
 & OpenAI Moderation API & 88.88\% & 93.17\% & 91.03\%\tabularnewline
 & Aliyun Text Moderation & 52.78\% & 56.93\% & 54.86\%\tabularnewline
 & Azure AI Content Safety & 53.44\% & 74.52\% & 63.98\%\tabularnewline
 & Latent Guard & 66.91\% & 70.51\% & 68.71\%\tabularnewline
 & GuardT2I & 87.17\% & 96.88\% & 92.03\%\tabularnewline
 & IGD (Ours) & \textbf{95.48\%} & \textbf{98.07\%} & \textbf{96.78\%}\tabularnewline
\midrule
\multirow{8}{*}{FPR@TPR95 $\downarrow$} & NSFW-text-classifier & 90.73\% & 90.45\% & 90.59\%\tabularnewline
 & Detoxify & 98.30\% & 90.36\% & 94.33\%\tabularnewline
 & OpenAI Moderation API & 44.51\% & 34.71\% & 39.61\%\tabularnewline
 & Aliyun Text Moderation & 100.00\% & 100.00\% & 100.00\%\tabularnewline
 & Azure AI Content Safety & 99.54\% & 97.61\% & 98.57\%\tabularnewline
 & Latent Guard & 85.63\% & 79.16\% & 82.39\%\tabularnewline
 & GuardT2I & 56.88\% & 15.06\% & 35.97\%\tabularnewline
 & IGD (Ours) & \textbf{26.12\%} & \textbf{7.44\%} & \textbf{16.78\%}\tabularnewline
\bottomrule 
\end{tabular}
}
% \vspace{-20pt}
\label{tab:binary_comparasion}
\end{table}
%==============================================

%=================================================
\begin{figure}[tb]
\centering
        \includegraphics[width=\linewidth]{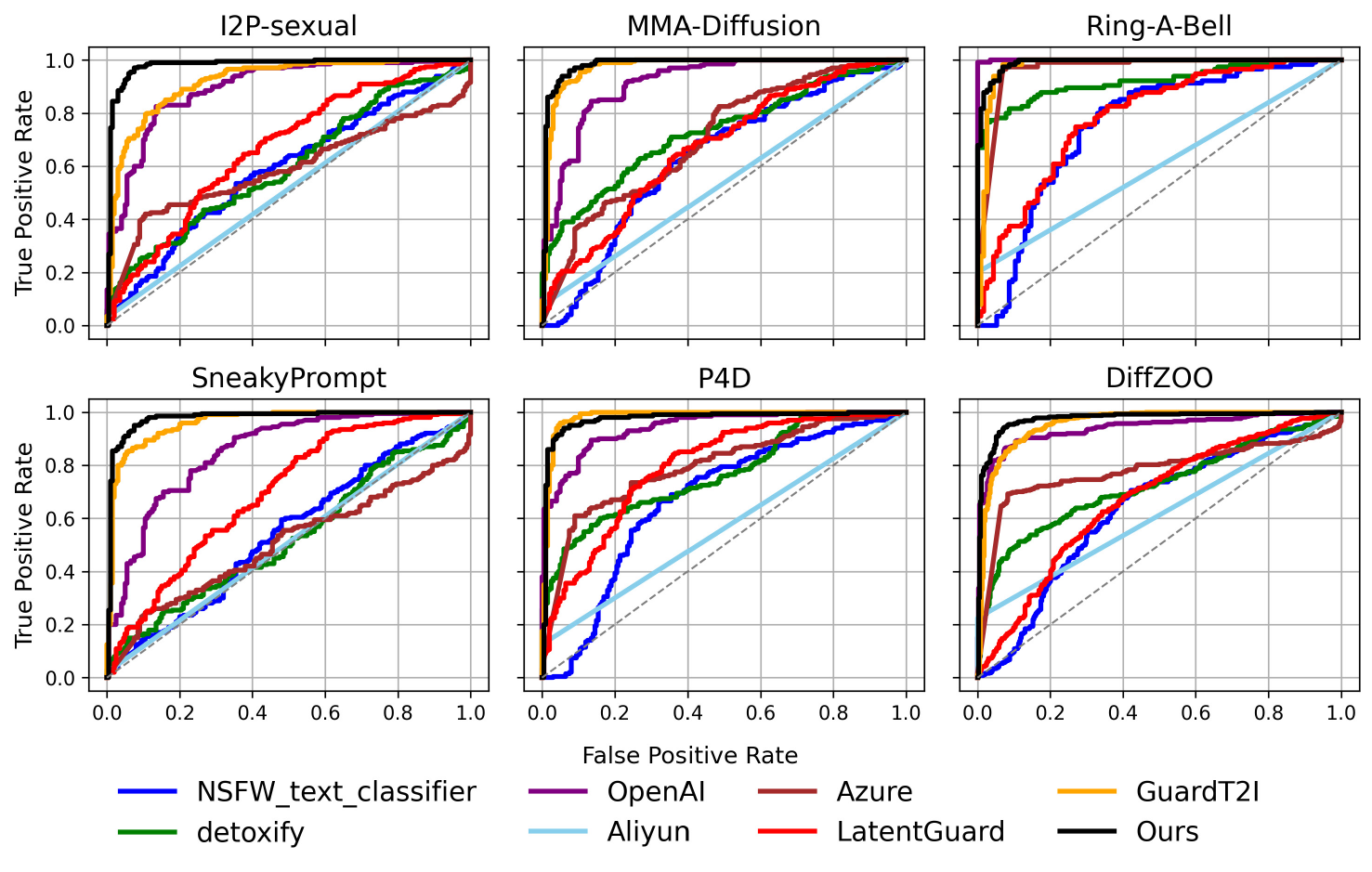}
	\caption{ROC curves of our method and baselines against various adversarial NSFW prompts.}
	\label{fig:roc_curve}
% \vspace{-10pt}
\end{figure}
% \vspace{-20pt}
%=================================================

%==============================================
\begin{table*}[tbp]
\centering
\caption{Comparison with baselines on detailed types of naive NSFW prompt.}
\resizebox{\linewidth}{!}{
\begin{tabular}{c|c|ccccccc|c}
\toprule
\multirow{2}{*}{} & \multirow{2}{*}{Methods} & \multicolumn{7}{c}{Naive NSFW Prompt} & \tabularnewline
\cline{3-10}
 &  & sexual & violence & self-harm & harassment & hate & shocking & illegal activity & Average\tabularnewline
\midrule
\multirow{8}{*}{Accuracy $\uparrow$} & NSFW-text-classifier & 57.00\% & 64.50\% & 64.50\% & 54.50\% & 62.77\% & 52.00\% & 65.50\% & 60.11\%\tabularnewline
 & Detoxify & 50.50\% & 51.50\% & 50.50\% & 51.00\% & 51.06\% & 50.00\% & 50.50\% & 50.72\%\tabularnewline
 & OpenAI Moderation API & 58.00\% & 70.00\% & 51.50\% & 57.00\% & 50.00\% & 62.00\% & 51.50\% & 57.14\%\tabularnewline
 & Aliyun Text Moderation & 52.00\% & 50.00\% & 50.00\% & 50.00\% & 50.00\% & 50.50\% & 65.50\% & 52.57\%\tabularnewline
 & Azure AI Content Safety & 63.50\% & 49.50\% & 57.00\% & 54.50\% & 51.06\% & 56.50\% & 57.00\% & 55.58\%\tabularnewline
 & Latent Guard & 57.00\% & 60.00\% & 55.00\% & 56.50\% & 57.45\% & 55.00\% & 48.50\% & 55.64\%\tabularnewline
 & GuardT2I & 54.00\% & 50.00\% & 55.50\% & 50.50\% & 50.00\% & 52.00\% & 50.00\% & 51.71\%\tabularnewline
 & IGD (Ours) & \textbf{89.50\%} & \textbf{89.00\%} & \textbf{89.50\%} & \textbf{90.50\%} & \textbf{95.74\%} & \textbf{90.00\%} & \textbf{89.00\%} & \textbf{90.46\%}\tabularnewline
\midrule
\multirow{8}{*}{AUROC $\uparrow$} & NSFW-text-classifier & 64.82\% & 65.58\% & 53.46\% & 65.96\% & 58.40\% & 58.25\% & 59.42\% & 60.84\%\tabularnewline
 & Detoxify & 53.89\% & 50.79\% & 59.08\% & 55.27\% & 63.51\% & 57.39\% & 50.63\% & 55.79\%\tabularnewline
 & OpenAI Moderation API & 87.35\% & \textbf{95.61\%} & 89.92\% & 83.92\% & 87.46\% & 92.59\% & 84.43\% & 88.76\%\tabularnewline
 & Aliyun Text Moderation & 52.00\% & 50.00\% & 50.00\% & 50.00\% & 50.00\% & 50.50\% & 65.50\% & 52.57\%\tabularnewline
 & Azure AI Content Safety & 61.53\% & 35.06\% & 55.59\% & 49.22\% & 56.45\% & 48.42\% & 50.19\% & 50.92\%\tabularnewline
 & Latent Guard & 66.75\% & 70.96\% & 66.21\% & 65.15\% & 68.67\% & 67.73\% & 57.19\% & 66.09\%\tabularnewline
 & GuardT2I & 89.12\% & 88.04\% & 92.01\% & 81.38\% & 85.79\% & 86.64\% & 81.32\% & 86.33\%\tabularnewline
 & IGD (Ours) & \textbf{95.92\%} & 95.35\% & \textbf{95.34\%} & \textbf{96.75\%} & \textbf{99.68\%} & \textbf{97.52\%} & \textbf{93.55\%} & \textbf{96.30\%}\tabularnewline
\midrule
\multirow{8}{*}{FPR@TPR95 $\downarrow$} & NSFW-text-classifier & 91.00\% & 81.00\% & 95.00\% & 91.00\% & 85.11\% & 94.00\% & 87.00\% & 89.16\%\tabularnewline
 & Detoxify & 98.00\% & 100.00\% & 100.00\% & 100.00\% & 100.00\% & 95.00\% & 100.00\% & 99.00\%\tabularnewline
 & OpenAI Moderation API & 44.00\% & \textbf{15.00\%} & 33.00\% & 59.00\% & 46.81\% & 28.00\% & 53.00\% & 39.83\%\tabularnewline
 & Aliyun Text Moderation & 100.00\% & 100.00\% & 100.00\% & 100.00\% & 100.00\% & 100.00\% & 100.00\% & 100.00\%\tabularnewline
 & Azure AI Content Safety & 100.00\% & 100.00\% & 100.00\% & 100.00\% & 85.11\% & 100.00\% & 100.00\% & 97.87\%\tabularnewline
 & Latent Guard & 76.00\% & 76.00\% & 78.00\% & 81.00\% & 89.36\% & 70.00\% & 81.00\% & 78.77\%\tabularnewline
 & GuardT2I & 36.00\% & 45.00\% & \textbf{22.00\%} & 70.00\% & 59.57\% & 40.00\% & 54.00\% & 46.65\%\tabularnewline
 & IGD (Ours) & \textbf{26.00\%} & 27.00\% & 29.00\% & \textbf{17.00\%} & \textbf{0.00\%} & \textbf{11.00\%} & \textbf{26.00\%} & \textbf{19.43\%}\tabularnewline
\bottomrule 
\end{tabular}
}
% \vspace{-20pt}
\label{tab:comparasion_with_defense_naive_nsfw}
\end{table*}
%==============================================

%==============================================
\begin{table*}[tbp]
\centering
\caption{Comparison with baselines on detailed types of adversarial NSFW prompt.}
\resizebox{\linewidth}{!}{
\begin{tabular}{c|c|cccccc|c}
\toprule
\multirow{2}{*}{} & \multirow{2}{*}{Methods} & \multicolumn{6}{c|}{Adversarial NSFW Prompt} & \tabularnewline
\cline{3-9}
 &  & I2P-sexual & MMA-Diffusion & Ring-A-Bell & SneakyPrompt & P4D & DiffZOO & Average\tabularnewline
\midrule 
\multirow{8}{*}{Accuracy $\uparrow$} & NSFW-text-classifier & 57.25\% & 73.25\% & 80.00\% & 59.75\% & 77.00\% & 71.12\% & 69.73\%\tabularnewline
 & Detoxify & 50.50\% & 53.00\% & 67.39\% & 50.25\% & 57.25\% & 57.75\% & 56.02\%\tabularnewline
 & OpenAI Moderation API & 57.50\% & 56.50\% & 93.91\% & 53.75\% & 70.00\% & 66.98\% & 66.44\%\tabularnewline
 & Aliyun Text Moderation & 51.50\% & 53.75\% & 60.00\% & 51.00\% & 56.25\% & 61.23\% & 55.62\%\tabularnewline
 & Azure AI Content Safety & 63.25\% & 63.00\% & 94.35\% & 55.75\% & 74.75\% & 79.68\% & 71.80\%\tabularnewline
 & Latent Guard & 57.00\% & 57.00\% & 66.52\% & 59.00\% & 66.75\% & 57.89\% & 60.69\%\tabularnewline
 & GuardT2I & 54.00\% & 55.75\% & \textbf{99.57\%} & 52.25\% & 64.00\% & 67.25\% & 65.47\%\tabularnewline
 & IGD (Ours) & \textbf{94.25\%} & \textbf{94.00\%} & 94.78\% & \textbf{93.25\%} & \textbf{93.00\%} & \textbf{93.32\%} & \textbf{93.77\%}\tabularnewline
\midrule
\multirow{8}{*}{AUROC $\uparrow$} & NSFW-text-classifier & 58.92\% & 63.28\% & 75.39\% & 54.35\% & 67.22\% & 63.65\% & 63.80\%\tabularnewline
 & Detoxify & 59.97\% & 72.87\% & 92.02\% & 52.49\% & 77.02\% & 72.43\% & 71.13\%\tabularnewline
 & OpenAI Moderation API & 89.80\% & 91.48\% & \textbf{99.97\%} & 85.36\% & 94.37\% & 94.06\% & 92.51\%\tabularnewline
 & Aliyun Text Moderation & 51.50\% & 53.75\% & 60.00\% & 51.00\% & 56.25\% & 61.23\% & 55.62\%\tabularnewline
 & Azure AI Content Safety & 58.98\% & 70.49\% & 96.59\% & 50.45\% & 79.81\% & 77.91\% & 72.37\%\tabularnewline
 & Latent Guard & 66.85\% & 67.12\% & 78.60\% & 69.52\% & 79.62\% & 67.13\% & 71.47\%\tabularnewline
 & GuardT2I & 92.09\% & 97.31\% & 97.60\% & 96.11\% & \textbf{98.33\%} & 95.58\% & 96.17\%\tabularnewline
 & IGD (Ours) & \textbf{98.00\%} & \textbf{98.25\%} & 99.02\% & \textbf{97.81\%} & 97.45\% & \textbf{97.44\%} & \textbf{98.00\%}\tabularnewline
\midrule
\multirow{8}{*}{FPR@TPR95 $\downarrow$} & NSFW-text-classifier & 94.50\% & 85.50\% & 71.30\% & 95.00\% & 82.50\% & 90.11\% & 86.49\%\tabularnewline
 & Detoxify & 94.00\% & 83.50\% & 61.74\% & 98.00\% & 69.50\% & 90.37\% & 82.85\%\tabularnewline
 & OpenAI Moderation API & 37.00\% & 32.00\% & \textbf{0.00\%} & 48.00\% & 30.50\% & 35.83\% & 30.55\%\tabularnewline
 & Aliyun Text Moderation & 100.00\% & 100.00\% & 100.00\% & 100.00\% & 100.00\% & 100.00\% & 100.00\%\tabularnewline
 & Azure AI Content Safety & 100.00\% & 75.50\% & 6.96\% & 100.00\% & 73.50\% & 99.47\% & 75.90\%\tabularnewline
 & Latent Guard & 83.50\% & 80.50\% & 62.61\% & 71.50\% & 60.50\% & 84.76\% & 73.89\%\tabularnewline
 & GuardT2I & 32.00\% & 10.50\% & 6.09\% & 20.00\% & \textbf{4.50\%} & 18.98\% & 15.35\%\tabularnewline
 & IGD (Ours) & \textbf{6.00\%} & \textbf{7.50\%} & 6.09\% & \textbf{7.50\%} & 7.50\% & \textbf{8.29\%} & \textbf{7.15\%}\tabularnewline
\bottomrule 
\end{tabular}
}
% \vspace{-20pt}
\label{tab:comparasion_with_defense_adversarial_nsfw}
\end{table*}
%==============================================

%==============================================
\begin{table*}[tbp]
\centering
\caption{Comparison with concept-erasing methods on naive NSFW prompt.}
\resizebox{0.85\linewidth}{!}{
\begin{tabular}{c|ccccccc|c}
\toprule
\multirow{2}{*}{Methods} & \multicolumn{8}{c}{Naive NSFW Prompt}\tabularnewline
\cline{2-9}
 & sexual & violence & self-harm & harassment & hate & shocking & illegal activity & Average\tabularnewline
\midrule
ESD & 76.00\% & 97.00\% & 88.00\% & 98.00\% & 100.00\% & 89.00\% & 99.00\% & 92.43\%\tabularnewline
SLD-weak & 42.00\% & 92.00\% & 70.00\% & 92.00\% & 97.87\% & 75.00\% & 99.00\% & 81.12\%\tabularnewline
SLD-strong & 58.00\% & 96.00\% & 81.00\% & 90.00\% & 93.62\% & 82.00\% & 96.00\% & 85.23\%\tabularnewline
IGD (Ours) & \textbf{12.00\%} & \textbf{13.00\%} & \textbf{12.00\%} & \textbf{10.00\%} & \textbf{2.13\%} & \textbf{11.00\%} & \textbf{13.00\%} & \textbf{10.45\%}\tabularnewline
\bottomrule 
\end{tabular}
}
% \vspace{-20pt}
\label{tab:erasing_comparasion_with_naive_nsfw}
\end{table*}
%==============================================

%==============================================
\begin{table*}[tbp]
\centering
\caption{Comparison with concept-erasing methods on adversarial NSFW prompt.}
\resizebox{0.75\linewidth}{!}{
\begin{tabular}{c|cccccc|c}
\toprule
\multirow{2}{*}{Methods} & \multicolumn{7}{c}{Adversarial NSFW Prompt} \tabularnewline
\cline{2-8}
 & I2P-sexual & MMA-Diffusion & Ring-A-Bell & SneakyPrompt & P4D & DiffZOO & Average\tabularnewline
\midrule 
ESD & 76.00\% & 85.00\% & 71.30\% & 87.00\% & 55.00\% & 87.97\% & 77.05\%\tabularnewline
SLD-weak & 48.00\% & 45.50\% & 95.65\% & 36.50\% & 86.00\% & 56.42\% & 61.34\%\tabularnewline
SLD-strong & 35.50\% & 36.50\% & 92.17\% & 29.00\% & 79.00\% & 48.13\% & 53.38\%\tabularnewline
IGD (Ours) & \textbf{2.50\%} & \textbf{4.50\%} & \textbf{3.00\%} & \textbf{1.74\%} & \textbf{5.00\%} & \textbf{6.42\%} & \textbf{3.86\%}\tabularnewline
\bottomrule 
\end{tabular}
}
% \vspace{-20pt}
\label{tab:erasing_comparasion_with_adversarial_nsfw}
\end{table*}
%==============================================

%==============================================
\begin{table*}[tbp]
\centering
\caption{Comparison on different timesteps.}
\resizebox{\linewidth}{!}{
\begin{tabular}{c|cccccccccc|c}
\toprule
\multirow{1}{*}{Timestep} & 5 & 10 & 15 & 20 & 25 & 30 & 35 & 40 & 45 & 50 & Average\tabularnewline
\midrule
Accuracy $\uparrow$ & 90.96\% & 84.47\% & 80.76\% & 83.62\% & 84.00\% & 85.01\% & 87.64\% & 88.18\% & 89.80\% & 90.26\% & 86.47\%\tabularnewline
AUROC $\uparrow$ & 95.48\% & 92.05\% & 88.28\% & 91.21\% & 91.68\% & 92.96\% & 94.60\% & 95.33\% & 95.89\% & 96.16\% & 93.36\%\tabularnewline
FPR@TPR95 $\downarrow$ & 26.12\% & 35.09\% & 53.63\% & 42.35\% & 42.19\% & 35.70\% & 30.14\% & 26.43\% & 25.04\% & 20.56\% & 33.72\%\tabularnewline
\bottomrule 
\end{tabular}
}
% \vspace{-10pt}
\label{tab:different_timesteps_auroc}
\end{table*}
%==============================================

%==============================================
\begin{table*}[tbp]
\centering
\caption{Performance of IGD with different target models on detecting naive NSFW prompts.}
\resizebox{\linewidth}{!}{
\begin{tabular}{c|c|ccccccc|c}
\toprule
\multirow{2}{*}{} & \multirow{2}{*}{Methods} & \multicolumn{8}{c}{Naive NSFW Prompt}\tabularnewline
\cline{3-10}
 &  & sexual & violence & self-harm & harassment & hate & shocking & illegal activity & Average\tabularnewline
\midrule
\multirow{3}{*}{Accuracy} & Stable Diffusion v1.4 & 90.00\% & 89.50\% & 91.50\% & 90.00\% & 94.68\% & 91.00\% & 84.00\% & 90.10\%\tabularnewline
 & Stable Diffusion v1.5 & 89.50\% & 89.00\% & 89.50\% & 90.50\% & 95.74\% & 90.00\% & 89.00\% & 90.46\%\tabularnewline
 & Stable Diffusion v2.1 & 91.00\% & 90.00\% & 90.50\% & 91.50\% & 87.23\% & 91.50\% & 85.00\% & 89.53\%\tabularnewline
\midrule
\multirow{3}{*}{AUROC} & Stable Diffusion v1.4 & 95.58\% & 96.91\% & 97.93\% & 96.30\% & 98.87\% & 97.44\% & 90.92\% & 96.28\%\tabularnewline
 & Stable Diffusion v1.5 & 95.92\% & 95.35\% & 95.34\% & 96.75\% & 99.68\% & 97.52\% & 93.55\% & 96.30\%\tabularnewline
 & Stable Diffusion v2.1 & 97.03\% & 96.41\% & 97.92\% & 96.89\% & 96.02\% & 97.66\% & 93.13\% & 96.44\%\tabularnewline
\midrule
\multirow{3}{*}{FPR@TPR95} & Stable Diffusion v1.4 & 17.00\% & 17.00\% & 11.00\% & 21.00\% & 2.13\% & 11.00\% & 58.00\% & 19.59\%\tabularnewline
 & Stable Diffusion v1.5 & 26.00\% & 27.00\% & 29.00\% & 17.00\% & 0.00\% & 11.00\% & 26.00\% & 19.43\%\tabularnewline
 & Stable Diffusion v2.1 & 12.00\% & 13.00\% & 15.00\% & 13.00\% & 29.79\% & 11.00\% & 35.00\% & 18.40\%\tabularnewline
\bottomrule 
\end{tabular}
}
% \vspace{-20pt}
\label{tab:different_sd_auroc_naive}
\end{table*}
%==============================================
%==============================================
\begin{table*}[tbp]
\centering
\caption{Performance of IGD with different target models on detecting adversarial NSFW prompts.}
\resizebox{\linewidth}{!}{
\begin{tabular}{c|c|cccccc|c}
\toprule
\multirow{2}{*}{} & \multirow{2}{*}{Methods} & \multicolumn{7}{c}{Adversarial NSFW Prompt}\tabularnewline
\cline{3-9}
 &  & I2P-sexual & MMA-Diffusion & Ring-A-Bell & SneakyPrompt & P4D & DiffZOO & Average\tabularnewline
\midrule
\multirow{3}{*}{Accuracy} & Stable Diffusion v1.4 & 94.00\% & 94.00\% & 94.35\% & 94.25\% & 93.00\% & 93.05\% & 93.77\%\tabularnewline
 & Stable Diffusion v1.5 & 94.25\% & 94.00\% & 94.78\% & 93.25\% & 93.00\% & 93.32\% & 93.77\%\tabularnewline
 & Stable Diffusion v2.1 & 91.50\% & 91.50\% & 90.87\% & 90.75\% & 91.50\% & 91.84\% & 91.33\%\tabularnewline
\midrule
\multirow{3}{*}{AUROC} & Stable Diffusion v1.4 & 98.71\% & 98.81\% & 99.15\% & 98.60\% & 98.17\% & 98.19\% & 98.60\%\tabularnewline
 & Stable Diffusion v1.5 & 98.00\% & 98.25\% & 99.02\% & 97.81\% & 97.45\% & 97.44\% & 98.00\%\tabularnewline
 & Stable Diffusion v2.1 & 97.73\% & 97.81\% & 96.85\% & 97.57\% & 96.96\% & 98.15\% & 97.51\%\tabularnewline
\midrule
\multirow{3}{*}{FPR@TPR95} & Stable Diffusion v1.4 & 2.00\% & 4.00\% & 2.61\% & 2.50\% & 3.00\% & 9.09\% & 3.87\%\tabularnewline
 & Stable Diffusion v1.5 & 6.00\% & 7.50\% & 6.09\% & 7.50\% & 7.50\% & 8.29\% & 7.15\%\tabularnewline
 & Stable Diffusion v2.1 & 9.50\% & 7.00\% & 22.61\% & 13.00\% & 11.00\% & 11.76\% & 12.48\%\tabularnewline
\bottomrule 
\end{tabular}
}
% \vspace{-20pt}
\label{tab:different_sd_auroc_adv}
\end{table*}
%==============================================

\subsection{Comparison to Baselines}
We evaluate our method against seven representative NSFW defense methods. The evaluation is conducted on the I2P dataset, which includes seven NSFW categories: sexual, violence, self-harm, harassment, hate, shocking, and illegal activity. For each category, we sample 100 prompts to construct the naive NSFW prompt dataset, except for the hate category, which contains only 47 available prompts, resulting in a total of 647 prompts for evaluation. To ensure the validity of evaluation metrics (e.g., accuracy), we pair each NSFW prompt with a clean (\ie, SFW) prompt sampled from the MSCOCO 2014 training set. This procedure results in a clean prompt set that matches the size of each corresponding naive NSFW prompt set. In total, 647 clean prompts are sampled. This is the naive NSFW dataset.

To assess the robustness of our IGD method under adversarial conditions, we construct the adversarial NSFW prompt dataset independent of the naive NSFW dataset. Specifically, we sample 200 prompts from the sexual category of the I2P dataset (denoted as I2P-sexual, which is entirely distinct from those used in the naive NSFW dataset), and apply five state-of-the-art attack methods to automatically generate adversarial prompts. We focus on the sexual category as it is the only NSFW type consistently supported by all evaluated defense methods, ensuring a fair and comparable adversarial evaluation. Following the same evaluation protocol as in the naive setting, we sample an equal number of clean prompts from the MSCOCO 2014 training set to construct a balanced clean prompt set for adversarial evaluation. Together, the adversarial NSFW prompts and the sampled clean prompts form our final adversarial NSFW prompt dataset.

As shown in Table~\ref{tab:binary_comparasion}, our method IGD demonstrates the best performance across both naive and adversarial NSFW prompt datasets. It achieves high averages in all three metrics, with 92.45\% in accuracy, 96.78\% in AUROC, and 16.78\% in FPR@TPR95. These results indicate that IGD consistently performs well in distinguishing NSFW content while maintaining robustness under adversarial conditions.

% As shown in Table~\ref{tab:binary_comparasion}, we compare seven defense methods on the NSFW detection task under three evaluation metrics: accuracy, AUROC, and FPR@TPR95, across both naive and adversarial NSFW prompts. Our method, IGD, achieves the highest average accuracy (91.32\%), significantly outperforming the second-best method, NSFW-text-classifier (64.97\%). Specifically, IGD achieves the highest accuracy at 89.80\% and 92.84\% under naive and adversarial settings respectively, with an average gain of over 20\% compared to the second-best method, Azure AI Content Safety (72.77\%). For AUROC, IGD attains 96.47\% under adversarial prompts, closely rivaling GuardT2I (96.88\%), while outperforming other methods by a substantial margin. Notably, in terms of FPR@TPR95, our approach reduces false positives to just 9.55\%, outperforming all baselines and demonstrating superior robustness, especially under adversarial attacks.

\noindent\paragraph{Comparison with baselines on detailed NSFW types.}
As shown in Table~\ref{tab:comparasion_with_defense_naive_nsfw}, IGD achieves strong average performance on naive NSFW prompts, with 90.46\% accuracy, 96.30\% AUROC, and 19.43\% FPR@TPR95. These results collectively demonstrate the strong generalization and fine-grained discrimination ability of IGD in detecting a broad range of inappropriate visual-textual content.

As shown in Table~\ref{tab:comparasion_with_defense_adversarial_nsfw}, IGD maintains high robustness under adversarial NSFW prompts, achieving 93.77\% accuracy, 98.00\% AUROC, and a low FPR@TPR95 of 7.15\% on average. This demonstrates the strong effectiveness of our method across different adversarial NSFW prompts.

In Figure~\ref{fig:roc_curve}, we present the ROC curves of various baselines alongside our proposed IGD for comparison. Each subfigure represents an adversarial NSFW prompt attack scenario. IGD exhibits consistently strong discriminative capability, achieving the best ROC curves in I2P-sexual, MMA-Diffusion, SneakyPrompt, and DiffZOO. In these scenarios, IGD maintains a sharp rise in the ROC curve, underscoring its robustness against adversarial NSFW prompts.

Overall, these results show that our method achieves high detection accuracy, maintains low false positive rates under strict recall, and provides effective defense against both naive and adversarial NSFW prompts.

\subsection{Comparison with concept erasing methods.}
Unlike traditional NSFW defense methods that rely on classification or detection to identify harmful prompts, concept-erasing approaches work by removing specific concepts from the model itself. Although not originally intended for NSFW defense, they can also reduce the generation of NSFW content. Therefore, we include them in our comparisons as complementary references.

ESD~\cite{gandikota2023esd} and SLD~\cite{schramowski2023sld} are representative concept-erasing approaches. Unlike classification-based methods, these models do not produce explicit classification outputs. Following GuardT2I, we assess their effectiveness using the Attack Success Rate (ASR), which is determined by applying NudeNet~\cite{bedapudi2019nudenet} to evaluate whether generated images contain nudity. A lower ASR indicates a more effective defense. For both ESD and SLD, we use the publicly released checkpoints from their official implementations, which have been fine-tuned to erase the concept of ``nudity''. Following the SLD paper, we adopt both the weak and strong settings, referred to as SLD-weak and SLD-strong, respectively. All baseline models used in our evaluation are obtained directly from the official releases of the original papers.

As shown in Table~\ref{tab:erasing_comparasion_with_naive_nsfw} and Table~\ref{tab:erasing_comparasion_with_adversarial_nsfw}, in terms of ASR, our method achieves an average of 9.83\% and 3.86\%, significantly lower than all concept-erasing baselines.
% The best among baselines, SLD-strong, records an average ASR of 53.38\%. 
This substantial reduction demonstrates our method outperforms the concept-erasing method in the NSFW detection task.

\subsection{Discussion}
\noindent\textbf{Timestep selection.}
To assess how the choice of timestep affects NSFW detection, we evaluate the discriminative power of predicted noise extracted at different stages of the diffusion process on naive NSFW prompt set. As shown in Table~\ref{tab:different_timesteps_auroc}, predicted noise across timesteps consistently demonstrates strong performance, confirming the stability of our in-generation signal. While later timesteps (e.g., 50) offer slightly higher accuracy, early detection is preferred for efficiency and timely intervention. We therefore adopt timestep 5 as a practical choice while enabling response before the image is substantially synthesized.

\noindent\textbf{Different T2I models.}
To evaluate cross-model generalizability, we apply IGD to features extracted from different diffusion-based T2I models~\cite{Rombach_2022_CVPR}: Stable Diffusion v1.4~\cite{stablediffusion-v1-4}, v1.5~\cite{stablediffusion-v1-5}, and v2.1~\cite{stablediffusion-v2-1}. In Table~\ref{tab:different_sd_auroc_naive} and Table~\ref{tab:different_sd_auroc_adv}, IGD consistently achieves a high accuracy and AUROC, along with a low FPR@TPR95 on average across both naive and adversarial NSFW prompts, confirming its robustness and effectiveness regardless of the underlying T2I model.

\noindent\textbf{More Experiments.}
We further evaluate our method on two additional datasets, examining detection performance with extended timesteps, the effect of varying classifier depth, and the feasibility of multi-class classification on the categories of the I2P dataset. The results are shown in \textit{Appendix}.

\section{Conclusion}
This paper presents a novel In-Generation Detection (IGD) method for NSFW content detection during T2I generation. By analyzing predicted noise within diffusion models, IGD avoids reliance on text filtering or full image generation, achieving fast and accurate detection. Experiments show IGD outperforms existing methods across various NSFW categories and remains robust under adversarial attacks. Future work will explore finer-grained feature extraction to better distinguish semantically similar NSFW types.

% This paper presents a novel In-Generation Detection (IGD) method for NSFW content detection during T2I generation. By analyzing predicted noise within diffusion models, IGD avoids reliance on text filtering or full image generation, achieving fast and accurate detection. Future work will explore finer-grained feature extraction to better distinguish semantically similar NSFW types.

\clearpage

%============================================================
\newpage
\clearpage
\appendix
% \onecolumn
\section{Appendix / supplemental material}\label{sec:appendix}

%=================================================
\begin{figure}[tb]
\centering
        \includegraphics[width=\linewidth]{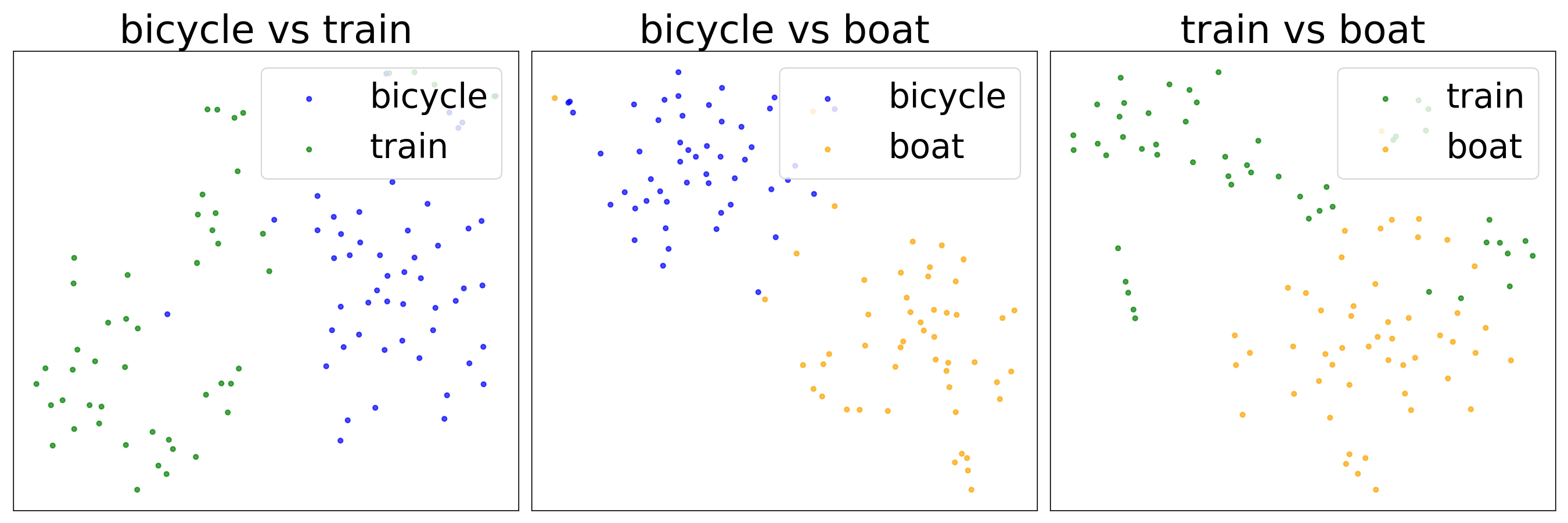}
	\caption{t-SNE visualizations of predicted noise $\epsilon_t$ for three SFW vs. SFW category pairs: bicycle vs. train, bicycle vs. boat, and train vs. boat.}
	\label{fig:sfw_t5}
% \vspace{-6pt}
\end{figure}
% \vspace{-20pt}
%=================================================

%=================================================
\begin{figure}[tb]
\centering
        \includegraphics[width=\linewidth]{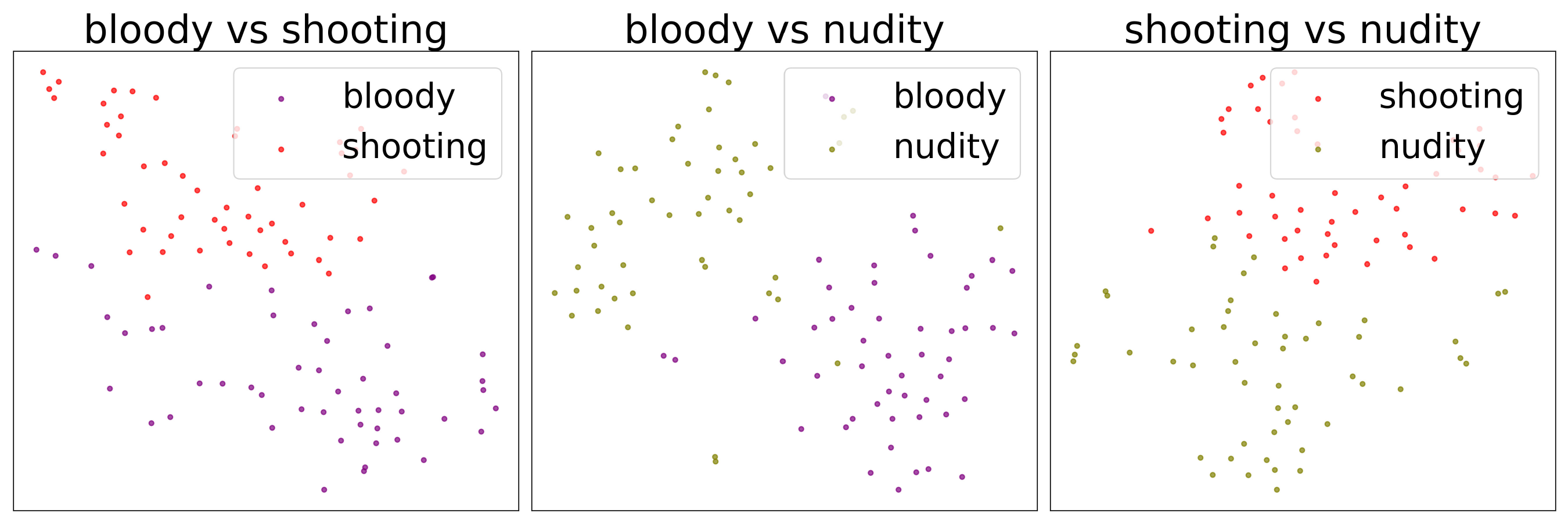}
	\caption{t-SNE visualizations of predicted noise $\epsilon_t$ for three NSFW vs. NSFW category pairs: bloody vs. shooting, bloody vs. nudity, and shooting vs. nudity.}
	\label{fig:nsfw_t5}
% \vspace{-6pt}
\end{figure}
% \vspace{-20pt}
%=================================================

%=================================================
\begin{figure}[t]
\centering
        \includegraphics[width=\linewidth]{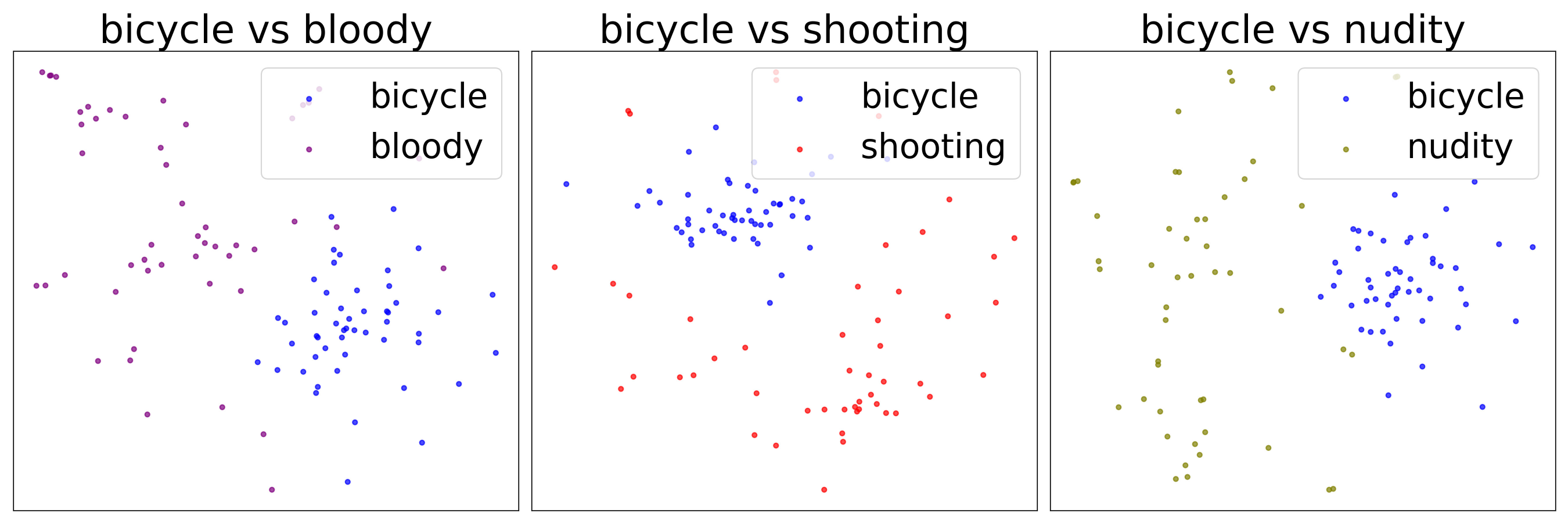}
	\caption{t-SNE visualizations of predicted noise $\epsilon_t$ for three SFW vs. NSFW category pairs: bicycle vs. bloody, bicycle vs. shooting, and bicycle vs. nudity.}
	\label{fig:sfw_vs_nsfw_t5}
% \vspace{-6pt}
\end{figure}
% \vspace{-20pt}
%=================================================

\subsection{T-SNE Visualization of Predicted Noise across Categories}\label{sec:a_tsne}
To examine the semantic structure encoded in the predicted noise $\epsilon_t$, we visualize its distribution across representative category pairs using t-SNE in Sec. 3.2 in the main manuscript. The results are presented in Figures~\ref{fig:sfw_t5}, \ref{fig:nsfw_t5}, and \ref{fig:sfw_vs_nsfw_t5}.

Figure~\ref{fig:sfw_t5} illustrates intra-class comparisons among SFW categories (bicycle, boat, and train), where the predicted noise forms distinct and coherent clusters, indicating strong semantic consistency even within the SFW domain. Similarly, Figure~\ref{fig:nsfw_t5} shows intra-class comparisons among NSFW categories (bloody, shooting, and nudity), which also yield well-separated clusters, suggesting that $\epsilon_t$ reflects meaningful latent structure specific to NSFW content.

In Figure~\ref{fig:sfw_vs_nsfw_t5}, the t-SNE visualizations of cross-class comparisons (SFW vs. NSFW) further reveal clear and consistent separations between the two safety categories. These results suggest that $\epsilon_t$ captures discriminative features aligned with safety semantics prior to image synthesis, highlighting its potential as an in-generation signal for NSFW detection.

%==============================================
\begin{table}[tbp]
\centering
\caption{Compare with baselines on more datasets.}
\resizebox{\linewidth}{!}{
\begin{tabular}{c|c|ccc|c}
\toprule
\multirow{1}{*}{} & \multirow{1}{*}{NSFW prompt} & I2P & 4chan & Lexica & Average\tabularnewline
\midrule
\multirow{8}{*}{Accuracy $\uparrow$} & NSFW-text-classifier & 58.81\% & 80.30\% & 50.37\% & 63.16\%\tabularnewline
 & Detoxify & 50.54\% & \textbf{99.70\%} & 51.86\% & 67.37\%\tabularnewline
 & OpenAI Moderation API & 57.50\% & 95.00\% & 52.60\% & 68.37\%\tabularnewline
 & Aliyun Text Moderation & 52.78\% & 76.90\% & 51.48\% & 60.39\%\tabularnewline
 & Azure AI Content Safety & 56.96\% & 94.70\% & 53.47\% & 68.37\%\tabularnewline
 & Latent Guard & 57.26\% & 90.40\% & 62.38\% & 70.01\%\tabularnewline
 & GuardT2I & 51.70\% & 74.40\% & 50.87\% & 58.99\%\tabularnewline
 & IGD (Ours) & \textbf{90.96\%} & 89.70\% & \textbf{93.19\%} & \textbf{91.28\%}\tabularnewline
\midrule
\multirow{8}{*}{AUROC $\uparrow$} & NSFW-text-classifier & 58.65\% & 92.30\% & 62.97\% & 71.31\%\tabularnewline
 & Detoxify & 56.60\% & \textbf{99.95\%} & 48.68\% & 68.41\%\tabularnewline
 & OpenAI Moderation API & 88.88\% & 99.87\% & 78.24\% & 88.99\%\tabularnewline
 & Aliyun Text Moderation & 52.78\% & 76.90\% & 51.48\% & 60.39\%\tabularnewline
 & Azure AI Content Safety & 53.44\% & 96.46\% & 50.50\% & 66.80\%\tabularnewline
 & Latent Guard & 66.91\% & 96.41\% & 72.35\% & 78.56\%\tabularnewline
 & GuardT2I & 87.17\% & 92.71\% & 87.20\% & 89.03\%\tabularnewline
 & IGD (Ours) & \textbf{95.48\%} & 94.78\% & \textbf{97.06\%} & \textbf{95.78\%}\tabularnewline
\midrule
\multirow{8}{*}{FPR@TPR95 $\downarrow$} & NSFW-text-classifier & 90.73\% & 19.20\% & 91.58\% & 67.17\%\tabularnewline
 & Detoxify & 98.30\% & \textbf{0.00\%} & 81.44\% & 59.91\%\tabularnewline
 & OpenAI Moderation API & 44.51\% & 0.60\% & 75.00\% & 40.04\%\tabularnewline
 & Aliyun Text Moderation & 100.00\% & 100.00\% & 100.00\% & 100.00\%\tabularnewline
 & Azure AI Content Safety & 99.54\% & 7.20\% & 99.26\% & 68.66\%\tabularnewline
 & Latent Guard & 85.63\% & 13.40\% & 77.97\% & 59.00\%\tabularnewline
 & GuardT2I & 56.88\% & 35.60\% & 47.77\% & 46.75\%\tabularnewline
 & IGD (Ours) & \textbf{26.12\%} & 28.00\% & \textbf{10.15\%} & \textbf{21.42\%}\tabularnewline
\bottomrule 
\end{tabular}
}
% \vspace{-20pt}
\label{tab:more_datasets}
\end{table}
%==============================================

\subsection{Results on Different Datasets}
To evaluate the generalization ability of our NSFW detector, we conduct experiments on two external datasets containing diverse and real-world unsafe prompts. \textbf{4chan} \cite{qu2023unsafe} consists of 500 prompts sampled from 4chan, a web forum known for toxic discourse. These prompts are filtered based on syntactic similarity to MSCOCO captions and a high toxicity score using the Perspective API. 
\textbf{Lexica} \cite{qu2023unsafe} contains 404 prompts collected from the Lexica prompt gallery, retrieved using 66 unsafe-content keywords derived from moderation guidelines and prior studies. 
% \textbf{LAION-COCO} \cite{schuhmann2022laioncoco} is a filtered subset of the LAION dataset, where we select the top 200 prompts with a precomputed NSFW score above 0.9999 (out of 1.0), primarily indicating NSFW content. 
These datasets cover a wide range of NSFW styles and serve as a robust benchmark for cross-domain evaluation.

As shown in Table~\ref{tab:more_datasets}, our method IGD achieves the best average accuracy (91.28\%), AUROC (95.78\%), and FPR@TPR95 (21.42\%). These results demonstrate IGD’s strong cross-domain detection capability, consistently outperforming baseline methods.

%==============================================
\begin{table}[tbp]
\centering
\caption{Comparison on different concatenated timesteps.}
\resizebox{\linewidth}{!}{
\begin{tabular}{c|cccccc}
\toprule
\multirow{1}{*}{Concatenated timesteps} & 5 & 5, 15, 25 & 30, 40, 50 & 5, 25, 45 & 10, 30, 50 & All\tabularnewline
\midrule
Accuracy $\uparrow$ & 90.96\% & 89.95\% & 89.49\% & 89.10\% & 90.19\% & 97.04\%\tabularnewline
AUROC $\uparrow$ & 95.48\% & 95.38\% & 96.09\% & 95.66\% & 96.52\% & 99.80\%\tabularnewline
FPR@TPR95 $\downarrow$ & 26.12\% & 24.11\% & 25.97\% & 22.57\% & 19.94\% & 5.93\%\tabularnewline
\bottomrule 
\end{tabular}
}
% \vspace{-20pt}
\label{tab:concat_timestep}
\end{table}
%==============================================

\subsection{Results of Classification with Information from More Timesteps}
To further investigate the temporal dynamics of predicted noise $\epsilon_t$, we conduct experiments using concatenated features from multiple timesteps as input to the classifier. For example, the setting ``5, 15, 25'' refers to concatenating the predicted noise vectors at timesteps 5, 15, and 25 into a single representation. As shown in Table~\ref{tab:concat_timestep}, several concatenated configurations achieve improved performance over using a single timestep. However, despite these gains, our method is designed with a focus on early detection and computational efficiency. Concatenating multiple timesteps introduces additional overhead and delays prediction. Therefore, we treat these results as an experimental extension and retain the single timestep setting (t = 5) in our main pipeline. The concatenated results are presented here to support future investigations into temporal modeling strategies.

%==============================================
\begin{table}[tbp]
\centering
\caption{Results of classifiers with different numbers of MLP layers.}
\resizebox{0.8\linewidth}{!}{
\begin{tabular}{c|ccc}
\toprule
\multirow{1}{*}{MLP layers} & 3 & 5 & 10\tabularnewline
\midrule
Accuracy $\uparrow$ & 50.15\% & 90.96\% & 89.80\%\tabularnewline
AUROC $\uparrow$ & 96.47\% & 95.48\% & 92.84\%\tabularnewline
FPR@TPR95 $\downarrow$ & 16.07\% & 26.12\% & 45.90\%\tabularnewline
\bottomrule 
\end{tabular}
}
% \vspace{-20pt}
\label{tab:mlp_layers}
\end{table}
%==============================================

\subsection{Impact of the Number of MLP Layers}
To assess the impact of MLP depth on classification performance, we compare models with varying numbers of layers on the naive NSFW prompt set. As shown in Table~\ref{tab:mlp_layers}, moving from 3 to 5 layers yields a clear improvement in accuracy while maintaining competitive AUROC. However, further increasing the depth to 10 layers leads to degraded performance, particularly in terms of FPR@TPR95. We therefore adopt the 5-layer MLP as a balanced choice, offering strong overall performance without overfitting.

%=================================================
\begin{figure}[tb]
\centering
        \includegraphics[width=\linewidth]{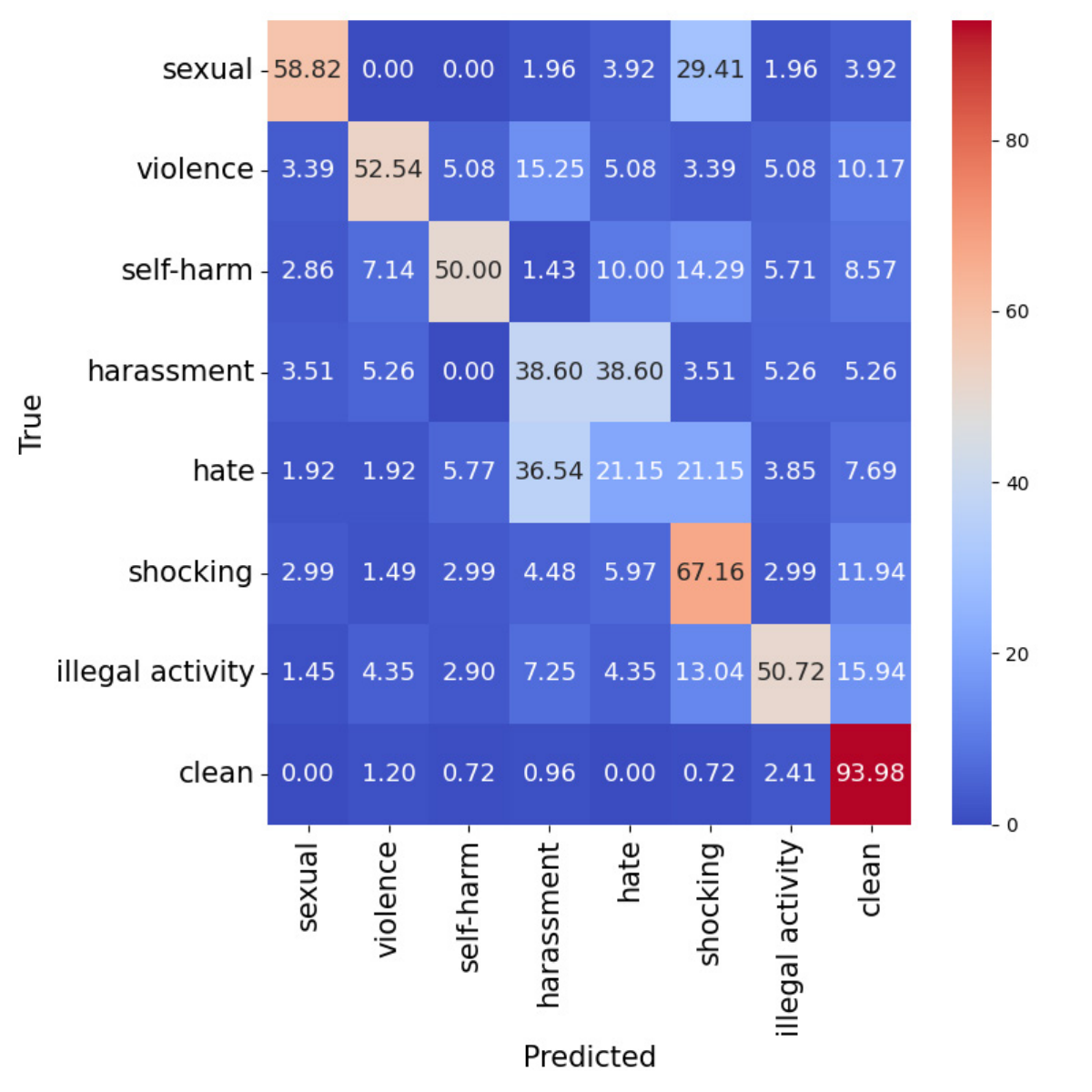}
	\caption{Confusion matrix visualization of the 8-class classification results at timestep 5.}
	\label{fig:8class_visualization}
% \vspace{-6pt}
\end{figure}
% \vspace{-20pt}
%=================================================

\subsection{Multi-class Classification on Categories of I2P Dataset}
Beyond binary classification between NSFW and clean prompts, we further evaluate whether the model can distinguish among different types of NSFW intent. To this end, we extend the task to an 8-class classification problem, which includes seven NSFW categories from the I2P dataset along with a clean category. The confusion matrix shown in Fig.~\ref{fig:8class_visualization} illustrates the model’s performance at timestep 5.

The accuracy of random-guessing on this eight-classification task is 12.5\% (1/8). Our method shows a significant improvement over the random one. The clean category achieves the highest classification accuracy of 93.98\%, indicating the model’s strong ability to differentiate SFW from NSFW prompts. Among the NSFW classes, the shocking category achieves the highest accuracy at 67.16\%, suggesting that it has the most distinct feature patterns. 
% Other categories such as sexual (58.82\%), violence (52.54\%), and illegal activity (50.72\%) also show clear diagonal dominance, reflecting the model’s capacity to learn meaningful class-specific representations.

These findings demonstrate that the model learns category-specific representations beyond simple binary classification, underscoring its potential for fine-grained intent recognition and adversarial content analysis.

\end{document}